\documentclass{article}
\usepackage{spconf,amsmath,graphicx}
\usepackage{booktabs}
\usepackage{url}
\usepackage[table]{xcolor}
\usepackage{amsfonts}
\DeclareMathOperator*{\argmin}{arg\,min}

\usepackage{enumitem}
\setlist{nosep, leftmargin=14pt}

\usepackage{mwe} 

\title{CRESTomics: Analyzing Carotid Plaques in the CREST-2 Trial with a New Additive Classification Model}
\name{\shortstack{Pranav Kulkarni$^{1,2}$, Brajesh K. Lal$^{1,3}$, Georges Jreij$^{3}$, Sai Vallamchetla$^{4}$, \protect\\ \textit{Langford Green}$^{3}$, \textit{Jenifer Voeks}$^{5}$, \textit{John Huston}$^{4}$, \textit{Lloyd Edwards}$^{6}$, \textit{George Howard}$^{6}$, \protect\\ \textit{Bradley A. Maron}$^{1,3}$, \textit{Thomas G. Brott}$^{4}$, \textit{James F. Meschia}$^{4}$, Florence X. Doo$^{1,3}$$^*$, \textit{Heng Huang}$^{1,2}$}$^*$}

\address{
$^{1}$University of Maryland Institute for Health Computing, $^{2}$University of Maryland, College Park \\
$^{3}$University of Maryland School of Medicine, $^{4}$Mayo Clinic, $^{5}$Medical University of South Carolina \\
$^{6}$University of Alabama at Birmingham
}

\begin{document}
%
\maketitle
\def\thefootnote{*}\footnotetext{Co-senior authors contributed equally to this work.}\def\thefootnote{\arabic{footnote}}
\begin{abstract}
Accurate characterization of carotid plaques is critical for stroke prevention in patients with carotid stenosis. We analyze 500 plaques from CREST-2, a multi-center clinical trial, to identify radiomics-based markers from B-mode ultrasound images linked with high-risk. We propose a new kernel-based additive model, combining coherence loss with group-sparse regularization for nonlinear classification. Group-wise additive effects of each feature group are visualized using partial dependence plots. Results indicate our method accurately and interpretably assesses plaques, revealing a strong association between plaque texture and clinical risk.
\end{abstract}
\section{Introduction} \label{sec:intro}

Carotid stenosis is a major risk factor for ischemic stroke, and its early identification and characterization is critical for stroke prevention \cite{saba2019imaging}. Conventional assessment of carotid plaques relies on duplex ultrasound (US) to measure luminal narrowing and plaque morphology (shape, tissue composition, and stability) \cite{brinjikji2015ultrasound}. Prior studies have shown that these characteristics can be measured reliably using histologically correlated imaging markers \cite{lal2024carotid,lal2002pixel,lal2006noninvasive}. While these markers provide valuable information about morphology, they are limited to a small set of features derived from pixel intensities.

Radiomics enables the extraction of high-dimensional imaging features from medical images, capturing subtle variations in texture, shape, and composition within a region-of-interest (ROI). Markers derived from radiomics have been correlated with treatment response, prognosis, and future outcomes across several medical domains \cite{parmar2015machine,bera2022predicting}. In carotid stenosis, US-based radiomics has been used to predict plaques with high-risk for cerebrovascular events \cite{hou2023radiomics,huang2024radiomics}. While DL methods could extract similar information, they are not viable due to the limited sample size of carotid plaque datasets.

In this paper, we analyze 500 plaques from Carotid Revascularization and Medical Management for Asymptomatic Carotid Stenosis Study (CREST-2), a multi-center randomized clinical trial investigating stroke prevention strategies for asymptomatic carotid stenosis \cite{howard2017carotid}. Our goal is to identify radiomics-based markers linked with high clinical risk from B-mode images. Existing radiomics ML methods either lack the flexibility to model nonlinear relationships or tradeoff interpretability for performance. Additive models offer a promising solution, where feature contributions ``add up" towards the prediction, but formulations like GAM \cite{hastie1986generalized} struggle with nonlinear dependencies and shrinking irrelevant feature contributions. We propose a new kernel-based additive model, combining coherence loss with group-sparsity for nonlinear classification while being fully interpretable.

\section{Methods} \label{sec:methods}

\begin{figure}[!t]
    \centering
    \includegraphics[width=1\linewidth]{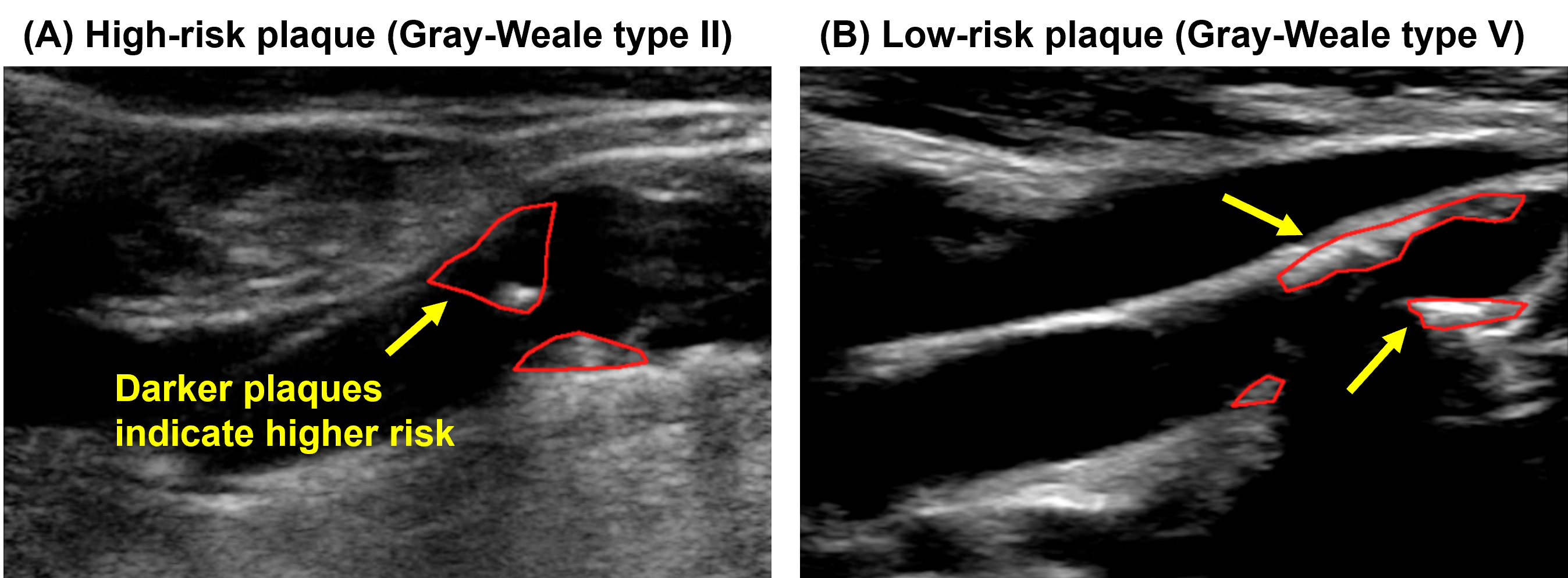}
    \caption{High- and low-risk carotid plaque on B-mode image.}
    \label{fig:example}
\end{figure}

\subsection{CREST-2}

We analyze carotid plaques of $n=500$ asymptomatic patients with $\geq70\%$ stenosis from CREST-2 \cite{howard2017carotid}. Each patient contains a normalized and segmented B-mode image capturing the plaque at the carotid bifurcation (Fig. \ref{fig:example}). Measurements include: 1) histologically correlated imaging features, comprising plaque grayscale median, longitudinal area (mm$^2$), and composition (mm$^2$) of hemorrhage, lipid, fibrous, muscle, and calcified tissue; 2) Gray-Weale classification \cite{gray1988carotid}; and 3) Hemodynamic features, comprising peak systolic velocity (PSV), end-diastolic velocity (EDV), and the internal carotid artery-to-common carotid artery (ICA/CCA) ratio. More details are provided in prior publications \cite{lal2024carotid}.

\subsection{Additive Classification Model}

Consider a binary classification problem with input space $\mathcal{X} \subset \mathbb{R}^p$ and labels $\mathcal{Y} \in \{-1, 1\}$. Given training samples $\mathbf{z}=\{(x_i, y_i)\}^n_{i=1}$ drawn from distribution $\rho$ on $\mathcal{X} \times \mathcal{Y}$, the goal is to learn $f:\mathcal{X} \to \mathbb{R}$ such that $\text{sgn}(f(x))$ approximates a Bayes classifier. 

Additive models \cite{hastie1986generalized} decompose this input space $\mathcal{X} = (\mathcal{X}_1, ..., \mathcal{X}_p)$ into a sum of univariate components: 
\begin{equation}
    f(x) = \sum_{j=1}^{p} f_j(x_j), \quad f_j \in \mathcal{F}_j
\end{equation}
where $\mathcal{F}_j$ is a set of smooth functions on $\mathcal{X}_j$. Additive classification models \cite{christmann2012consistency} are then formulated as the regularized empirical risk minimization problem:
\begin{equation}
    \hat{f} = \argmin_{f \in \mathcal{F}} \frac{1}{n} \sum_{i=1}^{n} \ell(y_i f(x)_i) + \lambda\Omega(f)
\end{equation}
where $\ell$ is a convex surrogate loss (e.g., hinge, logistic loss), $\Omega(f)$ is penalty on $f$, and $\lambda > 0$ is a regularization parameter.

Suppose $d$ groups are derived from $\{1,...,p\}$. Let $\mathcal{X}^{(j)}$ be input component space for $j$-th group, $1\leq j \leq d$,  and $f^{(j)}:\mathcal{X}^{(j)} \to \mathbb{R}^{(j)}$ is its corresponding component function. For training samples $\mathbf{z}=\{(x_i, y_i)\}^n_{i=1}$, the hypothesis space for kernel-based additive models is defined as:
\begin{equation}
    \mathcal{H}_\mathbf{z} = \big\{ f : f(x) = \sum_{j=1}^{d} \sum_{i=1}^{n} \alpha_i^{(j)} K^{(j)}(x_i^{(j)}, \cdot), \quad \alpha_i^{(j)} \in \mathbb{R} \big\}
\end{equation}
and its corresponding component space is defined as:
\begin{equation}
    \mathcal{H}^{(j)}_\mathbf{z} = \big\{ f^{(j)} : \sum_{i=1}^{n} \alpha_i^{(j)} K^{(j)}(x_i^{(j)}, \cdot), \quad \alpha_i^{(j)} \in \mathbb{R} \big\}
\end{equation}
where $K^{(j)}: \mathcal{X}^{(j)} \times \mathcal{X}^{(j)} \to \mathbb{R}$ is continuous bounded kernel. 

Coherence loss \cite{zhang2012coherence} is a Fisher-consistent, smooth and convex surrogate loss defined, along with its empirical risk:
\begin{align}
    \ell_\sigma(y, f(x)) &= \frac{\log(1 + e^{(1 - y f(x))/\sigma}}{\log(1 + e^{1/\sigma})}, \quad \sigma > 0 \\
    \mathcal{E}_\mathbf{z}(f) &= \frac{1}{n} \sum_{i=1}^{n} \ell_\sigma(y_i, f(x_i))
\end{align}

Our new kernel-based additive model is then formulated using coherence loss and group-sparse regularization as:
\begin{equation}
    \hat{f} = \argmin_{f\in\mathcal{H}_\mathbf{z}} \mathcal{E}_\mathbf{z}(f) + \lambda \Omega(f), \quad \Omega(f) = \inf \sum_{j=1}^d w_j \|\alpha^{(j)}\|_2
\end{equation}
where $\alpha^{(j)}=(\alpha^{(j)}_1, ..., \alpha^{(j)}_n)^T$ and $w_j>0$ are tuning parameters for each group.

Let $\mathbf{K}^{(j)}_i = ((K^{(j)})(x^{(j)}_1, x^{(j)}_i), ..., (K^{(j)})(x^{(j)}_n, x^{(j)}_i))^T \in \mathbb{R}^{n}$ and $\mathbf{K}_i = ((\mathbf{K}^{(1)}_i)^T, ..., (\mathbf{K}^{(d)}_i)^T)^T \in \mathbb{R}^{dn}$, the vectorized optimization reduces to:
\begin{equation}
    \hat{\alpha} = \argmin_{a \in \mathbb{R}^{dn}} \frac{1}{n} \sum_{i=1}^{n} \frac{\log(1 + e^{(1- \alpha^TK_i)/\sigma})}{\log(1 + e^{1/\sigma})} + \lambda \sum_{j=1}^d w_j\|\alpha^{(j)}\|_2
\end{equation}
and the final additive classifier is defined as:
\begin{equation}
    \hat{f}_\mathbf{z}(x) = \sum_{j=1}^d\sum_{i=1}^n \hat{\alpha}^{(j)}_{\mathbf{z},i} K^{(j)}(x^{(j)}_i, \cdot)
\end{equation}
Function contribution (i.e., importance) for $j$-th group is:
\begin{equation}
    \|\hat{f}^{(j)}\|_2 = \big\| \sum_{i=1}^{n} \hat{\alpha}_i^{(j)} K^{(j)}(x_i^{(j)}, \cdot) \big\|_2
\end{equation}

\section{Experimental Results} \label{sec:results}

We extract 102 radiomics features using PyRadiomics \cite{van2017computational}: 9 shape, 18 first-order, and 75 texture features (24 GLCM, 14 GLDM, 16 GLRLM, 16 GLSZM, and 5 NGTDM). All images are preprocessed using CLAHE-Advanced \cite{triantafyllou2025effect} for robustness and rescaled to isotropic size for consistency.

\begin{figure*}[!ht]
    \centering
    \includegraphics[width=1\linewidth]{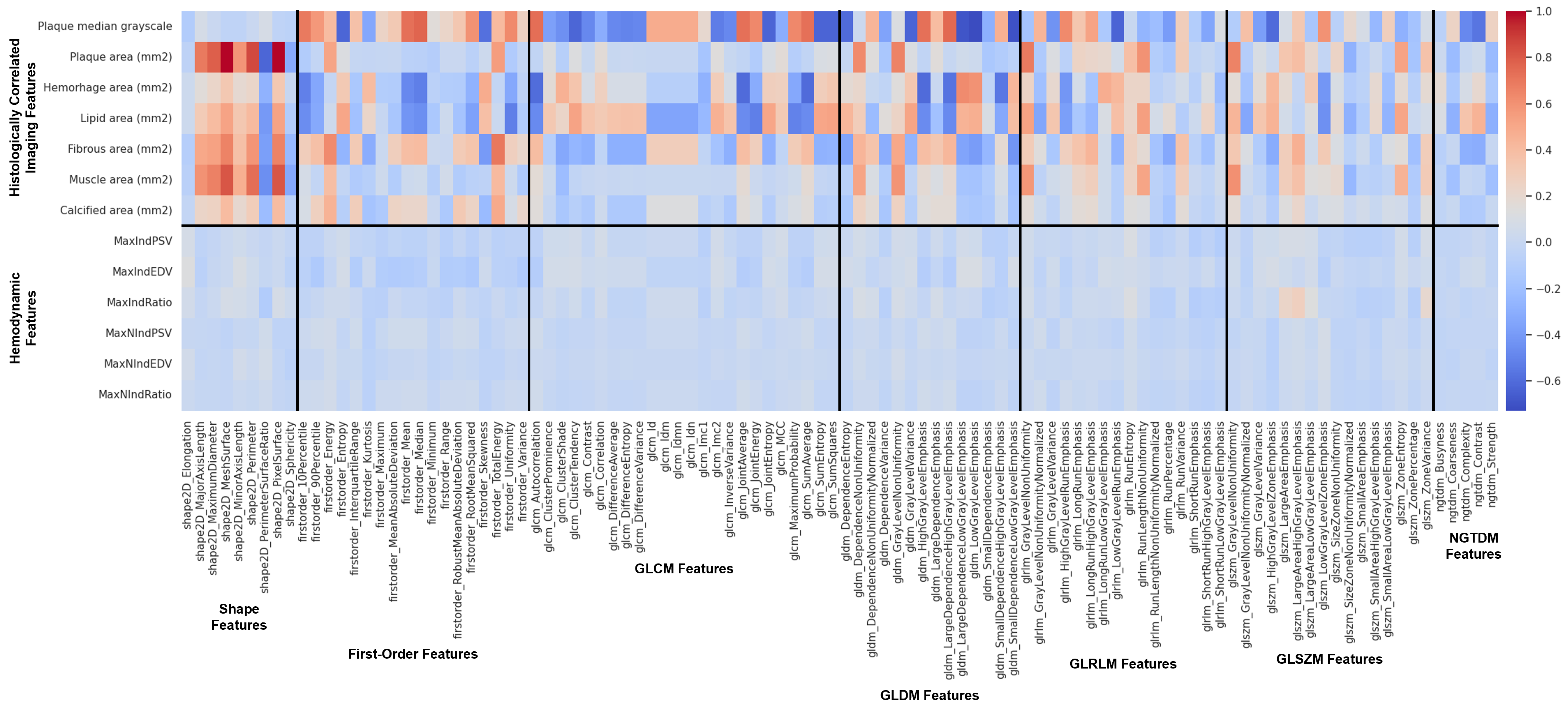}
    \caption{Pearson's $r$ between radiomics (x-axis) and plaque characteristics (y-axis).}
    \label{fig:correlation_summary}
\end{figure*}

\noindent\textbf{Correlation Analysis.} We measure Pearson's $r$ between radiomics and plaque characteristics (histologically correlated imaging markers and hemodynamic measurements) to determine whether radiomic features capture similar associations to established measurements and reveal additional relationships with plaque texture and shape. Results indicate that plaque grayscale median, hemorrhage area, lipid area, and fibrous area show moderate-to-strong correlations, while calcified area and hemodynamic measurements show weak correlations (Fig. \ref{fig:correlation_summary}). In contrast, plaque and muscle area are primarily associated with shape features.

\noindent\textbf{Clinical Risk Assessment.} We evaluate radiomics-based ML models to predict high-risk plaques, defined by Gray-Weale types I and II \cite{alexandratou2022advances} ($n=60$, $12\%$). To reduce dimensionality, top-10 radiomic features are selected via ElasticNet with 5-fold cross-validation and partitioned into groups based on their radiomic feature category.

\begin{table}[!ht]
\centering
\caption{Performance of classification models on CREST-2. Statistically significant differences are underlined.}
\label{tab:results}
\small
\begin{tabular}{lccc} \toprule \rowcolor{gray!25}
Method & AUROC$\uparrow$ & ACC$\uparrow$ & F1$\uparrow$ \\ \midrule
Logistic & \underline{0.93 (0.01)} & 96.00 (1.26) & \underline{79.52 (7.69)} \\
L1SVM & \underline{0.90 (0.02)} & 96.00 (1.26) & \underline{79.52 (7.69)} \\
L2SVM & \underline{0.90 (0.02)} & 96.00 (1.26) & \underline{79.52 (7.69)} \\
GaussianSVM & 0.94 (0.02) & \cellcolor{orange!15}\textbf{97.20 (0.98)} & 86.55 (5.34) \\
XGBoost & 0.93 (0.03) & 96.40 (0.80) & 82.18 (4.36) \\
GAM & \underline{0.92 (0.01)} & 96.80 (0.98) & 84.36 (5.34) \\ \midrule
\textbf{Ours} & \cellcolor{orange!15}\textbf{0.95 (0.01)} & \cellcolor{orange!15}\textbf{97.20 (1.60)} & \cellcolor{orange!15}\textbf{88.11 (5.59)} \\ \bottomrule
\end{tabular}
\end{table}

The model is trained on $n=450$ ($90\%$) samples with 5-fold cross-validation using Gaussian kernel $K(x_i, x_j) = \exp(-\gamma \|x_i - x_j\|^2)$ and optimized with groupwise majorization descent \cite{yang2015fast}. Class weights were specified to handle data imbalance and we set hyperparameters $\lambda=0.001$ and $\sigma=1$ via grid search. Baselines include logistic regression, linear SVM with L1 penalty, linear SVM with L2 penalty, nonlinear SVM with Gaussian kernel, XGBoost, and logistic GAM \cite{hastie1986generalized}. We measured AUROC, accuracy, and F1 score on a held-out test set ($n=50$, $10\%$). All values are reported as Mean (SD) and paired t-tests are used to test for statistical significance $p<0.05$. Our code is available at \url{https://github.com/itspranavk/CRESTomics}.

Results show that our model achieves best overall performance, with $0.95$ AUROC, $97.20\%$ accuracy, and $88.11\%$ F1 score (Table \ref{tab:results}). GaussianSVM and XGBoost perform comparably, but have limited interpretability due to their black-box nature. Linear models (Logistic, L1SVM, and L2SVM) underperform due to their inability to capture nonlinear relationships in data, and GAM’s lack of group-sparsity limits its ability to shrink irrelevant feature contributions.

\begin{figure}[!t]
    \centering
    \includegraphics[width=1\linewidth]{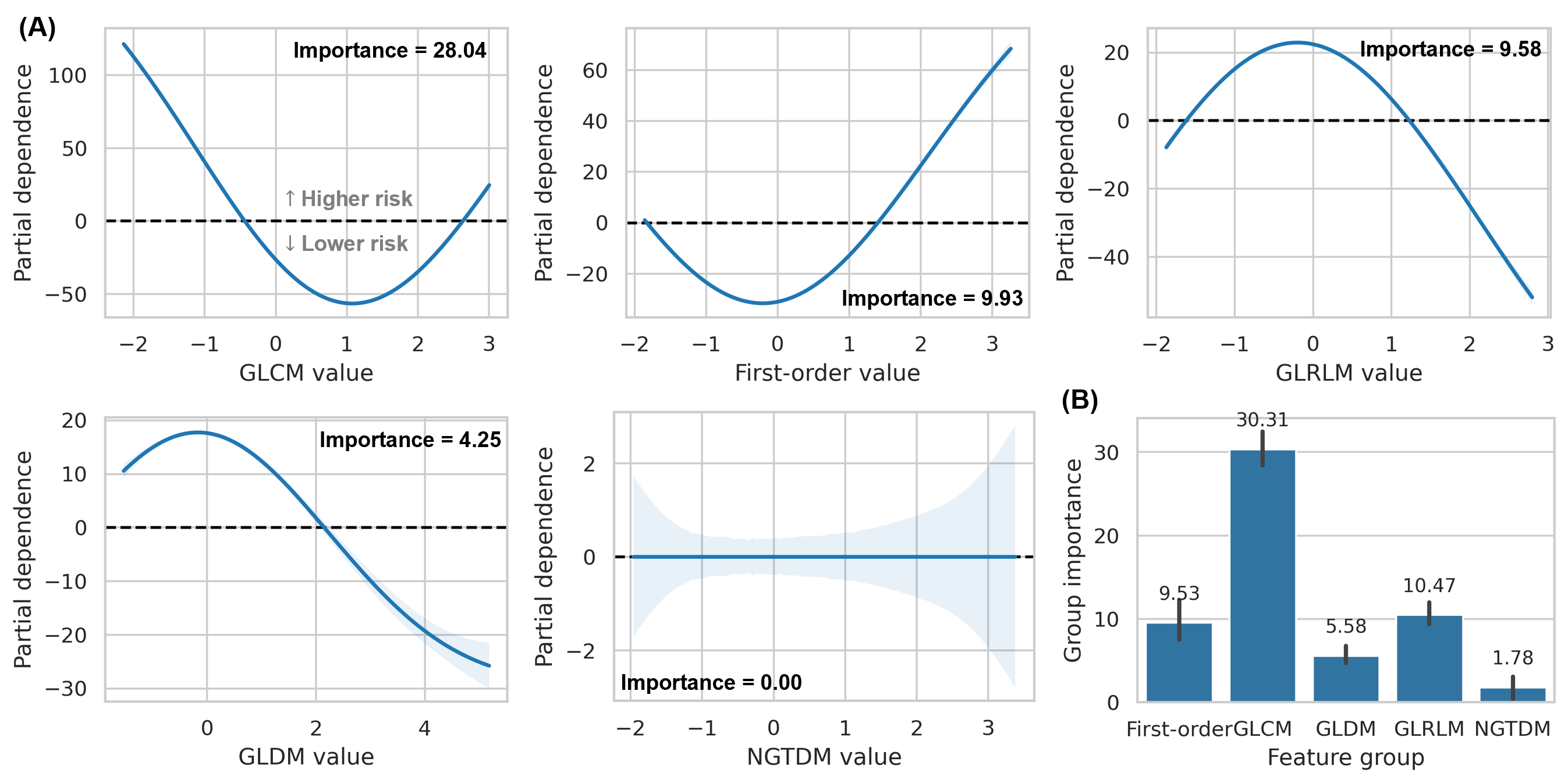}
    \caption{\textbf{(A)} Partial dependence plots for radiomic feature groups. Positive dependence indicates higher risk contributions, while negative dependence indicates lower risk contributions. \textbf{(B)} Group importance across 5-fold cross-validation.}
    \label{fig:pdep}
\end{figure}

\noindent\textbf{Interpretability.} We identify radiomics-based markers linked with plaque vulnerability using partial dependence plots and group contributions. Results indicate that GLCM texture features are most strongly associated with high-risk plaques, followed by first-order, GLRLM, and GLDM features (Fig. \ref{fig:pdep}). NGTDM texture features exhibit no apparent relationship based on their flat partial dependence curves.

\section{Conclusion} \label{sec:discussion}

Our analysis of carotid plaques from CREST-2, a multi-center randomized clinical trial, show that radiomic features are associated with histologically correlated imaging markers. However, hemodynamic measurements (like PSV) exhibit poor correlation as they depend on both plaque morphology and vessel obstruction, and thus, fail to capture luminal narrowing with plaque-only radiomic analysis.

Our proposed model accurately characterizes carotid plaques as either high- or low-risk and outperforms all baselines while being fully interpretable. Combining kernel-based additive structure with coherence loss and group-sparsity enables high nonlinear classification performance. Partial dependence allows feature contributions to be quantified and visualized. Results indicate that plaque texture features are strongly associated with clinical risk than first-order features, despite Gray-Weale type being based on plaque brightness. This aligns with conclusions from prior studies \cite{hou2023radiomics}. 

Limitations include high computational complexity of kernel-based methods and sensitivity of US-based radiomic analysis due to differences in acquisition and inter-rater variability in manual segmentation \cite{triantafyllou2025effect}. Future work includes linking radiomics-based markers with luminal narrowing to predict clinical outcomes for stroke prevention.

\section{Compliance with Ethical Standards} \label{sec:ethics}

CREST-2 protocol was approved by a Central IRB at the University of Cincinnati, and all participants provided written informed consent. ClinicalTrials.gov number, NCT02089217.

\section{Acknowledgments} \label{sec:acknowledgments}

Research was supported in part by the NIH NINDS under award number U01 NS080168 and U01 NS080165.

\bibliographystyle{IEEEbib}
\bibliography{refs}

\end{document}